\documentclass[runningheads]{llncs}
\usepackage{graphicx}
\usepackage{amsmath,amssymb,amsfonts}

\usepackage{xcolor}
\usepackage{booktabs}
\usepackage{multirow}
\usepackage{bbding}
\usepackage{flushend}
\begin{document}
\title{TreeNet: A lightweight One-Shot Aggregation Convolutional Network
}
%
%
\author{Lu Rao}
%
\authorrunning{Lu Rao}
%

%
\maketitle              
\begin{abstract}
The architecture of deep convolutional networks (CNNs) has evolved for years, becoming more accurate and faster. However, it is still challenging to design reasonable network structures that aim at obtaining the best accuracy under a limited computational budget. In this paper, we propose a Tree block, named after its appearance, which extends the One-Shot Aggregation (OSA) module while being more lightweight and flexible. Specifically, the Tree block replaces each of the $3\times3$ Conv layers in OSA into a stack of shallow residual block (SRB) and $1\times1$ Conv layer. The $1\times1$ Conv layer is responsible for dimension increasing and the SRB is fed into the next step. By doing this, when aggregating the same number of subsequent feature maps, the Tree block has a deeper network structure while having less model complexity. In addition, residual connection and efficient channel attention(ECA) is added to the Tree block to further improve the performance of the network. Based on the Tree block, we build efficient backbone models calling TreeNets. TreeNet has a similar network architecture to ResNet, making it flexible to replace ResNet in various computer vision frameworks. We comprehensively evaluate TreeNet on common-used benchmarks, including ImageNet-1k for classification, MS COCO for object detection, and instance segmentation. Experimental results demonstrate that TreeNet is more efficient and performs favorably against the current state-of-the-art backbone methods. 

\end{abstract}
\section{Introduction}
The architecture of deep convolutional neural networks (CNNs) has been evolved for years, becoming more accurate and faster. Starting from the milestone work of AlexNet~\cite{DBLP:conf/nips/KrizhevskySH12}, many studies are continuously investigated to further improve the performance of deep CNNs~\cite{DBLP:conf/cvpr/HeZRS16,DBLP:conf/cvpr/HuangLMW17,DBLP:conf/cvpr/LeeHLBP19,DBLP:conf/cvpr/LeeP20,qing2021sa}. One critical issue is designing more reasonable network structures that aim to obtain the best accuracy under a limited computation budget. 
\begin{figure}[htb]
	\centering
	\includegraphics[width=0.7\linewidth]{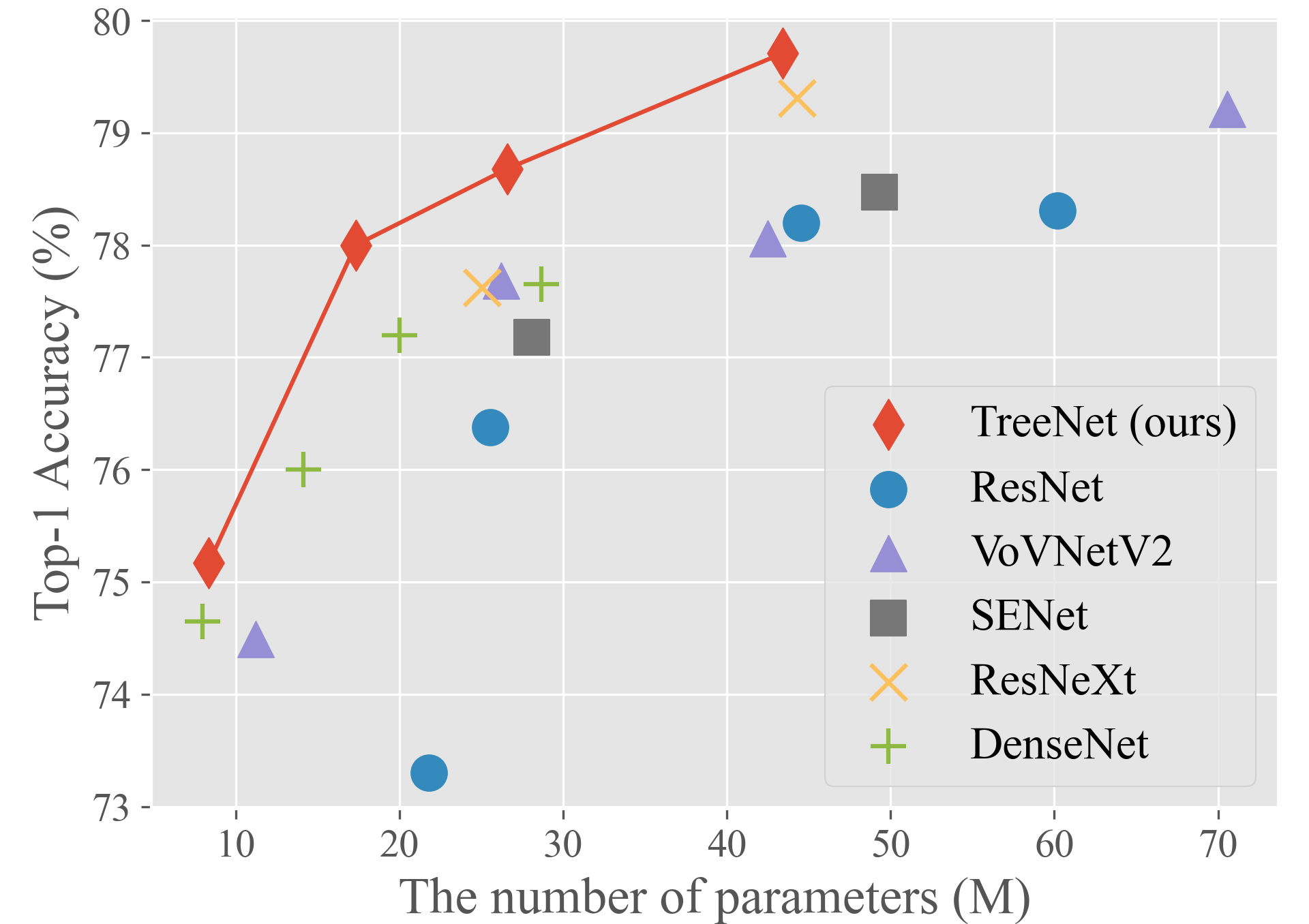}
	\caption{Comparisons of recently SOTA classification models on ImageNet-1k, including ResNet, ResNext, VoVNetV2, SENet, and DenseNet, in terms of accuracy and network parameters. Clearly, the proposed TreeNet achieves higher accuracy while having less model complexity.}
	\label{fig:fig1}
\end{figure}

ResNets~\cite{DBLP:conf/cvpr/HeZRS16} introduced the residual block which bypasses signal from one layer to the next via identity connections. DenseNets~\cite{DBLP:conf/cvpr/HuangLMW17} pointed out that when too many layers are employed in a residual block, it may impede the information flow in the network. To ensure maximum information flow between layers, DenseNets connected all layers (with matching feature-map sizes) directly with each other. However, the dense connections in intermediate layers are redundant, which induces inefficiencies. To improve the efficiency, VoVNet \cite{DBLP:conf/cvpr/LeeHLBP19} proposed a one-shot aggregation (OSA) block to concatenate all features only once in the last feature map, which makes input size constant and enables enlarging a new output channel. VoVNetV2~\cite{DBLP:conf/cvpr/LeeP20} further added the identity connection to aggregate the input and the output of OSA blocks.

\begin{figure*}[htb]
	\centering
	\includegraphics[width=1.0\linewidth]{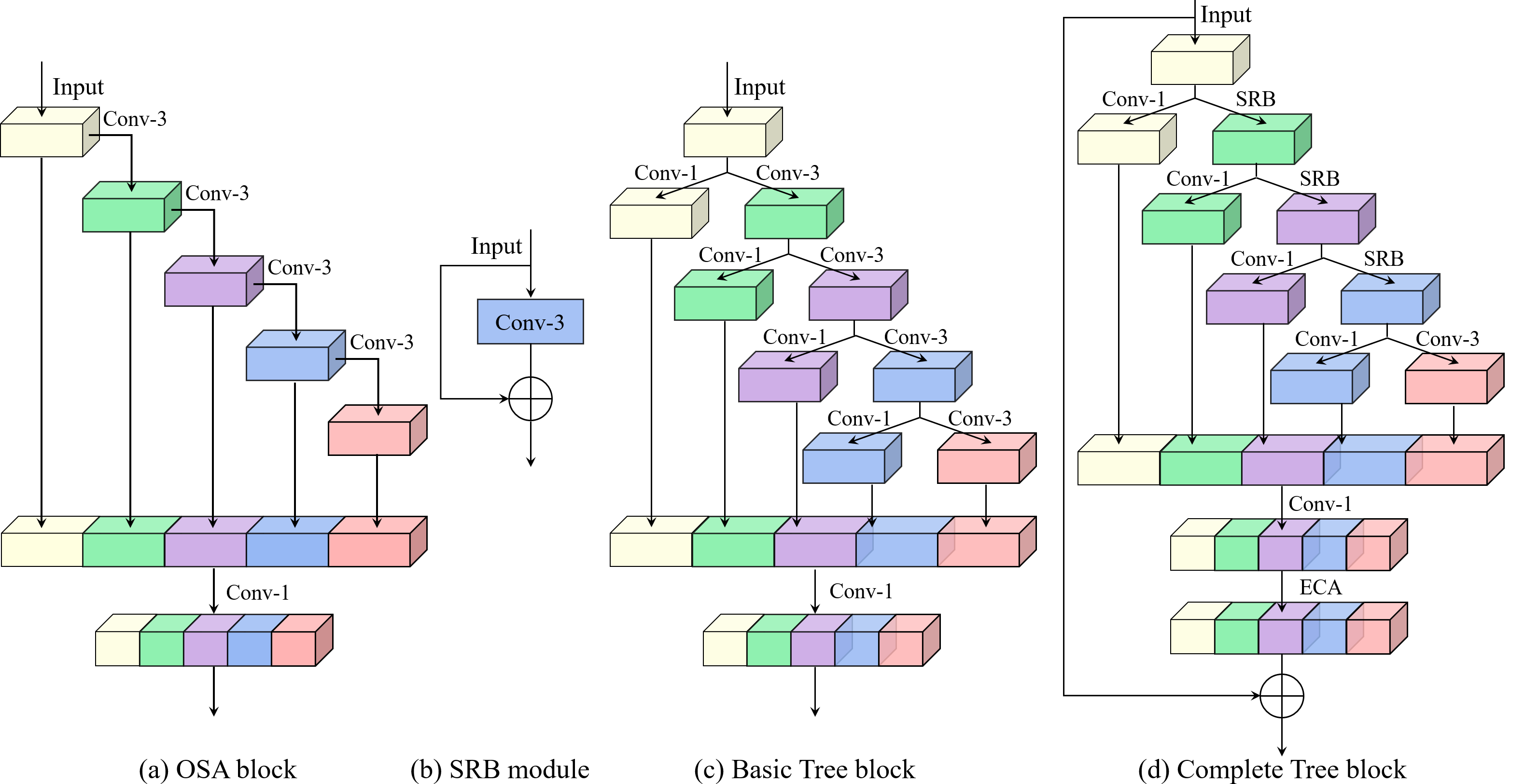}
	\caption{Comparison of OSA block and Tree block (with $\ell=4$ Conv-3 layers per block ). (a) OSA block. (b) SRB: shallow residual block. (c) Basic Tree block. (d) Complete Tree block: Tree block with residual connection and efficient channel attention (ECA). ``Conv-1", ``Conv-3" denote $1\times1$, $3 \times 3$ convolutional layer respectively. $\oplus$ denotes element-wise addition.}
	\label{fig:fig3}
\end{figure*}

Although OSA blocks show great potential in performance improvement, it is not lightweight enough. There is still room to improve efficiency. As illustrated in Fig.~\ref{fig:fig3}(a), the OSA block contains $\ell$ consecutive $3\times3$ Conv layers, each with $k$ filters,  after that, a transition layer is adopted to improve the model compactness. In OSA  block, $k$ tends to be high, which means the network's efficiency is highly affected by $k$. Therefore, reducing the size of $k$ while keeping the ability of feature extracting is an effective way to improve model efficiency.

Motivated by the bottleneck architecture introduced in ResNet, we propose a lightweight Tree block (named for its appearance), which modifies each of the OSA's $3\times3$ Conv layer(except the last one) into a half bottleneck structure (the bottom half). That is, each $3\times3$ Conv layer is replaced by a stack of a $3\times3$ layer (with fewer filters than $k$) and a $1\times1$ convolution. Where the $1\times1$ layer is responsible for increasing dimension, and the $3\times3$ layer is further processed by succeeding convolution layers. In addition, we add a $1\times1$ Conv layer (with $k$) filters to the input. The output of all $1\times1$ layers and the last $3\times3$ layer are then concatenated and then transformed by a transition layer(illustrated in Fig.~\ref{fig:fig3}(c)). As a result, when aggregating the same number of subsequent feature maps, the Tree block has a deeper network structure while having less model complexity.

Similar to VoVNetV2~\cite{DBLP:conf/cvpr/HuSS18}, Tree block can also benefit from residual connection and channel attention. Therefore, we add identity mapping and efficient channel attention (ECA~\cite{DBLP:conf/cvpr/WangWZLZH20}) to the Tree block. Compared with SE~\cite{DBLP:conf/cvpr/HuSS18}, ECA is more lightweight, which only introduces hands of parameters to obtain similar improvement.

However, only connecting the input path with the end of the Tree block is too coarse to maximize benefit from residual connections. To address this issue, we introduce more fine-grained residual learning into Tree block, i.e., replacing the $3\times3$ Conv layers with shallow residual blocks (SRB)~\cite{DBLP:journals/corr/abs-2009-11551}. As shown in Fig.~\ref{fig:fig3}(b), the SRB contains a convolution layer and identity connection, which can benefit from the residual learning without introducing extra parameters compared with plain convolutions. The complete Tree block is shown in Fig.~\ref{fig:fig3}(d).

\begin{figure*}[htb]
	\centering
	\includegraphics[width=1.0\linewidth]{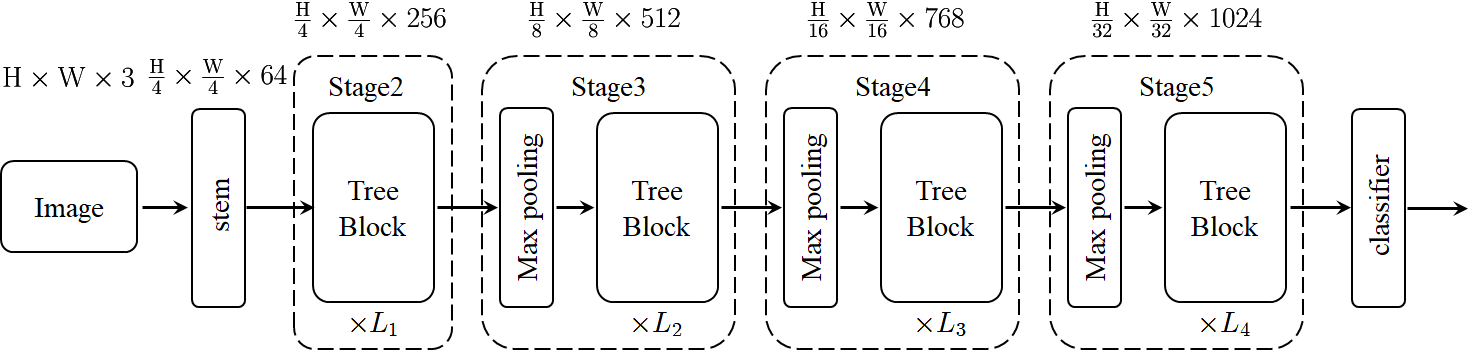}
	\caption{The pipeline of the proposed TreeNet. Similar to ResNet~\cite{DBLP:conf/cvpr/HeZRS16}, TreeNet building stages with stacked blocks, making it flexible to serve as the backbone of downstream tasks, such as Object detection, Person ReID, and Instance Segmentation, etc.}
	\label{fig:fig4}
\end{figure*}

Based on the Tree block, we build an efficient TreeNet (shown in Fig.~\ref{fig:fig4}), a similar network structure to ResNet~\cite{DBLP:conf/cvpr/HeZRS16} and SENet~\cite{DBLP:conf/cvpr/HuSS18}. We evaluate TreeNet on the ImageNet classification dataset. In particular, compared with the current state-of-the-art VoVNetV2 with the same number of layers, TreeNet can achieve higher accuracy (an average 0.6\% improvement) while having a fewer model complexity (illustrated in Fig.~\ref{fig:fig1}). We further evaluate TreeNet as backbones on the object detection and instance segmentation benchmarks, experimental results demonstrate that TreeNet can consistently achieve better accuracy than its competitors.

The contributions of this paper are summarized as follows: 
(1) We propose a lightweight yet effective deep convolutional structure Tree Block and based on which, we construct a memory-efficient backbone TreeNet. TreeNet is quite flexible, which can easily replace ResNet and served as the backbone for various computer vision tasks.
(2) We comprehensively evaluate TreeNet on ImageNet-1k and MS COCO 2017. Results demonstrate that TreeNet has less model complexity while achieving very competitive performance.

\section{Related Work}
\subsection{Efficient convolutional networks.}
Designing deep neural network architectures for the optimal trade-off between accuracy and efficiency has been an active research area in recent years~\cite{DBLP:conf/cvpr/HeZRS16,DBLP:conf/cvpr/HuangLMW17}. Here, we focus on the handcrafted structures since they tend to have better generalization ability. Among them, residual bottleneck structures are the most commonly used structures, which are shared by ResNet series~\cite{DBLP:conf/cvpr/HeZRS16}. Specifically, a residual bottleneck can be defined as a stack of one $1\times1$, several $3\times3$, and one $1\times1$ Conv layers with residual learning.
Recent works explore replacing the $3\times3$ Conv layers in the bottleneck with more complex modules or incorporation with attention modules~\cite{qing2021sa,DBLP:conf/cvpr/HuSS18}. Although achieving better accuracy, networks of these works are generally slower than the original version. Other works are dedicated to changing the connectivity structure of the internal convolutional blocks such as in ShuffleNet and DenseNet~\cite{DBLP:conf/cvpr/HuangLMW17}. These works usually concatenate all sub-features, therefore, they can also be called concatenation structures. Some works also show that concatenation structures can further boost with residual learning~\cite{DBLP:conf/cvpr/LeeP20}. Our TreeNet is one of these structures, i.e., a combination of concatenation with residual learning.


\subsection{Channel Attention.}
Channel attention mechanisms have been attracting increasing attention in research communities since it helps to improve the representation of interests. There are mainly two types of channel attention mechanisms most commonly used in deep CNNs: squeeze-and-excitation (SE)~\cite{DBLP:conf/cvpr/HuSS18} and efficient channel attention (ECA)~\cite{DBLP:conf/cvpr/WangWZLZH20}.  Specifically, for the given feature maps, SE first employs a global average pooling (GAP) for each channel independently, then two fully-connected (FC) layers with non-linearity followed by a Sigmoid function are used to generate channel weights. ECA simplifies the process of computing channel weights, i.e., after a channel-wise GAP without dimensionality reduction, ECA adopts a 1D convolution to capture local cross-channel interaction by considering every channel and its neighbors. Here, we directly embedded ECA into the Tree block.

\section{TreeNet}
In this section, we begin with revising the original OSA. We analyze its shortcomings and propose our solution, i.e., Tree block. Then introduce the details of constructing Tree Block. And finally, based on Tree block, we build TreeNet, an efficient backbone for dense prediction.

\subsection{Rethinking OSA Block}
As shown in Fig.~\ref{fig:fig3}(a), for a given feature map $\mathrm{x}_{0} \in \mathbb{R}^{c\times h\times w}$, where $c, h, w$ are channel dimension (i.e., number of filters), spatial height and width, respectively.
OSA block first adopts $\ell$ consecutive $3\times3$ Conv layers to capture the subsequent feature maps. Then these feature maps and the input are aggregated at once. Specifically, the output can be represent as 
\begin{equation}
	\mathrm{x} = [\mathrm{x}_{0},\mathrm{x}_{1},\cdots, \mathrm{x}_{\ell}]
\end{equation}
where $[\mathrm{x}_{0},\mathrm{x}_{1},\cdots, \mathrm{x}_{\ell}]$ refers to the concatenation of the feature maps produced in layer $0, \cdots, \ell$. $\mathrm{x}_{i}=H_{i}(\mathrm{x}_{i-1})$ is the output of layer $i$, and $H(\cdot)$ is a composite function of three consecutive operations: a $3\times3$ convolution (each with $k$ filters), followed by batch normalization (BN)~\cite{DBLP:conf/icml/IoffeS15} and a rectified linear unit(ReLU)~\cite{DBLP:journals/jmlr/GlorotBB11}. After that, a transition layer ($1\times1$ Conv layer) is applied to reduce the number of feature maps. 

Let $k_{in}$ be the channel dimension of the input feature map, $k_{cat}$ be the channel dimension of the transition layer, then the parameters (parameters in BN are neglected) of a OSA block will be
\begin{equation}
	P_{o} = 9\times k_{in}\times k + 9(\ell-1) k^2 + (k_{in} + k\times \ell) k_{cat}
\end{equation} 

If $k_{in}$, $\ell$ and $k_{cat}$ are fixed, then the total parameters $P_{o}$ is proportional to $k$. 

One way to reduce the number of parameters is to adopt smaller $k$, but it will weaken the feature extraction capabilities of OSA. Another way is increasing $k$ while applying group convolutions instead of standard ones, however, this multi-branch structure will inevitably reducing the inference speed of the network since it greatly increases the memory access cost (MAC). Here, we give an example to calculate the MAC of the latter method to reduce the number of parameters. 

To simplify, we set the increased filters to $\hat{k}$, which is 4 times of $k$ (similar to ResNeXt50~\cite{DBLP:conf/cvpr/XieGDTH17}), and group number is $g$. Formally, the MAC of a standard $3\times3$ Conv layer $M_{s}$ and the new group $3\times3$ Conv layer $M_{g}$ can be represent as
\begin{equation}
	\begin{split}
		M_{s} &= hw(c + k) + 9ck \\
		M_{g} &= hw(c + \hat{k}) + \frac{9c\hat{k}}{g}
	\end{split}
\end{equation}

Then, the MAC increment $M_i= M_g - M_s$ of group convolutional layers over standard ones is 
\begin{equation}
	\begin{split}
		M_i & = hw(\hat{k}- k) + \frac{9c\hat{k}}{g} - 9ck\\
		& = 3k(hw - 3c(1-\frac{4}{g}))
	\end{split}
	\label{eq:eq4}
\end{equation}
In most cases, $hw>3c$, which means $M_i>0$, that is, this group strategy will affect the inference speed of OSA.

\subsection{Basic Tree Block}
Motivated by the bottleneck structure, we propose basic Tree block, which is more lightweight and powerful than OSA. As shown in Fig.~\ref{fig:fig3}(c), each $3\times3$ Conv layer(except for the last one) is replaced by a stack of two Conv layers, the $1\times1$ Conv (with $k$ filters) is adopted to increasing channel dimension, leaving the $3\times3$ layer (with $k'$ filters, except the last one, which have $k$ filters) a bottleneck with smaller input/output dimensions. Also, we add a $1\times1$ Conv layer (with $k$ filters) to the input. The output of all $1\times1$ Conv layers and the last $3\times3$ layer are concatenated and then transformed by a transition layer(a $1\times1$ Conv layer with $k_{cat}$ filters). As a result, the basic Tree block has a deeper network structure than OSA.
The total parameters of a basic Tree block are
\begin{equation}
	P_{t}\! =\!k_{in}(k + 9k') + (\ell + 8)k\!\times\! k' + 9(\ell-2)k'^2 + (\ell+1)k\!\times\! k_{cat}
	\label{eq:eq2}
\end{equation} 

To simplify, we assume $k_{in}=k_{cat}=2k$, then
\begin{equation}
	P_{o}-P_{t} = 9(\ell +1)k^2 - (\ell+26)k\times k' - 9(\ell-2)k'^2 
	\label{eq:eq3}
\end{equation}
In OSA, $\ell=3$ or $\ell=5$, which means if $k'<17k/18$, then the Tree block is lightweight than OSA. In practice, we set $k' = max(128, k_{cat} /4)$, $k\in [128, 256, 512, 768]$, and $k_{cat} \in [256, 512, 768, 1024]$ to make a better trade-off.

\subsection{Residual Learning and Channel Attention}
It has widely demonstrated that residual connection can help the deep convolutional network optimization \cite{DBLP:conf/cvpr/HeZRS16}. Therefore, we also add the identity mapping to basic Tree Blocks. Specifically, the input path is connected to the end of a basic Tree block so that the gradients can be back-propagated in an end-to-end manner like ResNet. Besides, to better benefit from the residual connection, we replace each $3\times3$ Conv layer with an SRB module \cite{DBLP:journals/corr/abs-2009-11551}, which can produce fine-grained residual learning without introducing extra parameters compared with plain convolutions. 

In addition, the performance of Tree blocks can also be improved by correctly incorporation of channel attention mechanisms. Here, we embed an ECA module~\cite{DBLP:conf/cvpr/WangWZLZH20} into the basic Tree block. Specifically, given an input, ECA first employs a global average pool for each channel independently, then a fast 1D convolution is adopted to capture local cross-channel attention by considering every channel and its neighbors. Compared with SE~\cite{DBLP:conf/cvpr/HuSS18}, ECA is more lightweight. The complete Tree block is shown in Fig.~\ref{fig:fig3}(d). Unless explicitly stated, Tree block in following sections is the complete one.

\subsection{TreeNet}
Based on Tree block, we build efficient backbone models called TreeNets. TreeNet consists of a stem block and 4 stages of Tree blocks, the output stride of TreeNet is 32, the same as ResNet, making it flexible to replace ResNet on down-stream computer vision tasks. The stem block includes 3 Conv layers. At the beginning of each stage(except stage 2), feature maps are down-sampled by a $3\times3$ max pooling with stride 2. The exact network configurations applied on ImageNet are shown in Table~\ref{tab:tab1}.

\begin{table*}[htb]
	\caption{Architectures for ImageNet-1k. Concatenation layers per block (shown in Fig.~\ref{fig:fig3}) in ``20-layer" is $\ell=3$ and others $\ell=5$. Growth rate $k$ in ``stage2", ``stage3", ``stage4", ``stage5" are 128, 128, 192, 256 respectively.}
	\label{tab:tab1}
	\begin{center}
		\resizebox{1.0\columnwidth}{!}{
			\begin{tabular}{c|c|c|c|c|c}
				\toprule[1.2pt]
				Layer name & Output Size & 20-Layer & 40-Layer & 58-Layer & 100-Layer \\
				\midrule[1.1pt]
				stem & $56^2 \times 128$  & \multicolumn{4}{c}{$\left[               
					\begin{array}{ccc}
						3\times3, & 64, & \text{stride 2} \\ 
						3\times3, & 64, & \text{stride 1} \\ 
						3\times3, & \!\!\! 128, & \text{stride 2} \\
					\end{array}
					\right]$} \\
				\midrule
				stage2 & $56^2 \times 256$  & Tree block$\times1$  & Tree block$\times1$  & Tree block$\times1$  & Tree block$\times1$  \\
				
				\midrule
				\multirow{2}{*}{stage3} & \multirow{2}{*}{$28^2 \times 512$} & \multicolumn{4}{c}{max pool, $3\times3$, stride 2} \\
				\cmidrule{3-6}
				&  & Tree block$\times1$  & Tree block$\times1$  & Tree block$\times1$  & Tree block$\times3$  \\
				
				\midrule
				\multirow{2}{*}{stage4} & \multirow{2}{*}{$14^2 \times 768$} & \multicolumn{4}{c}{max pool, $3\times3$, stride 2} \\
				\cmidrule{3-6}
				& & Tree block$\times1$ &	Tree block$\times2$ &	Tree block$\times4$ &	Tree block$\times9$  \\ 
				
				\midrule
				\multirow{2}{*}{stage5} & \multirow{2}{*}{$7^2 \times 1024$} & \multicolumn{4}{c}{max pool, $3\times3$, stride 2} \\
				\cmidrule{3-6}
				&  & Tree block$\times1$ &	Tree block$\times2$ &	Tree block$\times3$ &	Tree block$\times3$  \\
				\midrule
				
				Classifier & $1^2 \times1000$ & \multicolumn{4}{c}{$7\times7$ global average pool, 1000D fully-connected, softmax} \\
				\midrule
				\multicolumn{2}{c|}{GFLOPs} & 4.20 & 6.68 &	7.91 &	13.24  \\
				\bottomrule[1.2pt]
		\end{tabular}}
	\end{center}
\end{table*}

\section{Experiments}
In this section, we firstly utilize ablation studies to validate the effectiveness of the Tree block. Then, we compare TreeNet with other state-of-the-art attention modules on various computer vision tasks (image classification, object detection, and instance segmentation) to evaluate its performance.

\subsection{Image Classification on ImageNet-1k}
{\bf Settings.}
For image classification, we benchmark the proposed TreeNet on ImageNet-1k, which contains 1.28M training images and 50k validation images from 1,000 classes.
We adopt exactly the same data augmentation and hyper-parameters settings in ResNet \cite{DBLP:conf/cvpr/HeZRS16}. Specifically, the input images are randomly cropped to $224 \times 224$ with random horizontal flipping. All the architectures are trained from scratch by SGD with weight decay 1e-4, momentum 0.9, and mini-batch size 256 (using 8 GPUs with 32 images per GPU) for 100 epochs, starting from the initial learning rate of 0.1 (with a linear warm-up of 5 epochs) and decreasing it by a factor of 10 every 30 epochs. For the testing on the validation set, the shorter side of an input image is first resized to 256, and a center crop of $224 \times 224$ is used for evaluation. To train VoVNetV2, we first add the same classification layer (details are shown in Table~\ref{tab:tab1}) as TreeNet. Therefore, it has the same number of layers as its TreeNet counterpart (e.g., VoVNetV2-19 actually has 20 layers).
All the models are implemented using the Pytorch toolkit\footnote{https://github.com/pytorch/pytorch}.

\begin{table}[htp]
	\caption{Ablation studies results of TreeNet-20 on ImageNet-1k in terms of Top-1/Top-5 accuracy(in \%).}
	\label{tab:tab2}
	\setlength{\tabcolsep}{4.0pt}{
		\begin{center}
			\begin{tabular}{cccccc}
				\toprule[1.2pt]
				w/ SRB & w/ Residual & w/ ECA & Top-1 Acc(\%) & Top-5 Acc(\%) \\ 
				\midrule[1.2pt]
				&  &  & 73.09 & 91.24 \\
				\midrule 
				\Checkmark &  &  &73.22 (\textcolor[rgb]{ 0,  0,  1}{+0.13})   & 91.32 \\
				\midrule
				\Checkmark & \Checkmark &  & 74.46 (\textcolor[rgb]{ 0,  0,  1}{+1.24}) & 92.00 \\ 
				\midrule
				\Checkmark & \Checkmark & \Checkmark & 75.17 (\textcolor[rgb]{ 0,  0,  1}{+0.71}) & 92.34 \\ 
				\bottomrule[1.2pt]
			\end{tabular}
		\end{center}
	}
\end{table}

{\bf Ablation Study.} We report the ablation studies of TreeNet-20 on ImageNet-1k, to thoroughly investigate the components of the Tree blocks. Table~\ref{tab:tab2} shows the results of gradually adding SRB~\cite{DBLP:journals/corr/abs-2009-11551}, Residual connection~\cite{DBLP:conf/cvpr/HeZRS16}, and ECA~\cite{DBLP:conf/cvpr/WangWZLZH20}.

As shown in Table~\ref{tab:tab2}, the SRB improves the performance by 0.13\% in terms of Top-1 accuracy. Since SRB introduces zero extra parameters compared with plain convolutions, it is worth embedding SRB into the Tree block. By adding Residual connection and ECA, the performance of the base method increase by a large margin. The Top-1 accuracy improvements are 1.24\% and 0.71\%, respectively, which can demonstrate the effectiveness of residual connection and ECA. We also adopt SE~\cite{DBLP:conf/cvpr/HuSS18} to replace ECA to test its effect on TreeNet-20, which will get 75.13\% in terms of Top-1 accuracy, similar to ECA~\cite{DBLP:conf/cvpr/WangWZLZH20}. Considering SE brings more parameters, it is not worth embedding into the Tree Block.

\begin{table*}[htb]
	\caption{Comparisons of different classification methods on ImageNet-1k in terms of network parameters, GFLOPs, Inference FPS, and Top-1/Top-5 accuracy (in \%). The best records and changes over VoVNetV2 are marked in \textbf{bold} and \textcolor[rgb]{ 0,  0,  1}{blue}, respectively.}
	\label{tab:tab3}
	\begin{center}
		\resizebox{1.0\columnwidth}{!}{
			\begin{tabular}{cccccc}
				\toprule[1.2pt]
				Methods & Param. (M) & GFLOPs & Inference FPS &Top-1 Acc (\%) & Top-5 Acc (\%) \\
				\midrule[1.1pt]
				
				ResNet-18 & 11.69  & 1.82 & 2632 & 69.76  & 89.08  \\
				\midrule
				VoVNetV2-19 & 11.21  & 4.14 & 1351  & 74.49  & 92.14  \\
				\midrule
				\textbf{TreeNet-20} & 8.37 (\textcolor[rgb]{ 0,  0,  1}{$\downarrow 2.84$})  & 4.20 (\textcolor[rgb]{ 0,  0,  1}{$\uparrow 0.06$})  & 1290 (\textcolor[rgb]{ 0,  0,  1}{$\downarrow 61$}) & \textbf{75.17} (\textcolor[rgb]{ 0,  0,  1}{$\uparrow0.68$})  & \textbf{92.34} (\textcolor[rgb]{ 0,  0,  1}{$\uparrow0.2$})  \\
				\midrule[1.1pt]
				
				ResNet-34 & 21.80  & 3.68  & 1786 & 73.30  & 91.42  \\
				\midrule
				VoVNetV2-39 & 26.21  & 7.10  & 943 & 77.58  & 93.65  \\
				\midrule
				\textbf{TreeNet-40} & 17.33 (\textcolor[rgb]{ 0,  0,  1}{$\downarrow 8.88$})  & 6.68 (\textcolor[rgb]{ 0,  0,  1}{$\downarrow 0.42$}) & 904 (\textcolor[rgb]{ 0,  0,  1}{$\downarrow 39$}) & \textbf{78.00} (\textcolor[rgb]{ 0,  0,  1}{$\uparrow0.42$})  & \textbf{93.95} (\textcolor[rgb]{ 0,  0,  1}{$\uparrow0.3$})  \\
				\midrule[1.1pt]

				ResNeXt-50 & 25.03 & 4.27 & 943 & 77.62 & 93.70 \\
				\midrule		
				ResNet-50 & 25.56  & 4.12 & 1020 & 76.38  & 92.91  \\
				\midrule
				SENet-50 & 28.09  & 4.13  & 943 & 77.18  & 93.67  \\
				\midrule					
				VoVNetV2-57 & 42.48  & 8.96 & 781 & 78.06  & 94.01  \\
				\midrule
				\textbf{TreeNet-58} & 26.58 (\textcolor[rgb]{ 0,  0,  1}{$\downarrow 15.90$})  & 7.93 (\textcolor[rgb]{ 0,  0,  1}{$\downarrow 1.03$}) & 753 (\textcolor[rgb]{ 0,  0,  1}{$\downarrow 28$}) & \textbf{78.78} (\textcolor[rgb]{ 0,  0,  1}{$\uparrow0.72$})  & \textbf{94.43} (\textcolor[rgb]{ 0,  0,  1}{$\uparrow0.42$})  \\
				\midrule[1.1pt]
				
				ResNeXt-101 & 44.30 & 7.99 & 435 & 79.31 & 94.53 \\
				\midrule				
				ResNet-101 & 44.55  & 7.85 & 667 & 78.20  & 93.91  \\
				\midrule		
				SENet-101 & 49.33  & 7.86 & 602 & 78.47  & 94.10  \\
				\midrule
				VoVNetV2-99 & 70.55  & 16.53 & 459 & 79.21  & 94.52  \\
				\midrule
				\textbf{TreeNet-100} & 43.42 (\textcolor[rgb]{ 0,  0,  1}{$\downarrow 27.13$})  & 13.24 (\textcolor[rgb]{ 0,  0,  1}{$\downarrow 3.29$}) & 476 (\textcolor[rgb]{ 0,  0,  1}{$\uparrow17$}) & \textbf{79.91} (\textcolor[rgb]{ 0,  0,  1}{$\uparrow0.70$})  & \textbf{94.86} (\textcolor[rgb]{ 0,  0,  1}{$\uparrow0.34$})  \\
				\bottomrule[1.2pt]											
			\end{tabular}
		}
	\end{center}
\end{table*}

{\bf Comparisons with state-of-the-art Backbone Methods.} We compare the proposed TreeNet with the current state-of-the-art backbone methods, including ResNet \cite{DBLP:conf/cvpr/HeZRS16}, ResNeXt~\cite{DBLP:conf/cvpr/XieGDTH17}, SENet~\cite{DBLP:conf/cvpr/HuSS18}, and VoVNetV2 \cite{DBLP:conf/cvpr/LeeP20}. The evaluation metrics include both efficiency (i.e., network parameters, GFLOPs and Inference FPS\footnote{Evaluate on one NVIDIA Tesla V100 GPU card.}) and effectiveness (i.e., Top-1/Top-5 accuracy). Results are shown in Table \ref{tab:tab3}. Here, we make several analysis:

(1) There is no doubt that SENet achieves better accuracy results than ResNet. Since the only difference of SENet and ResNet is channel attention, this can demonstrate that correctly incorporating attention mechanisms into convolution blocks can significantly improve the networks' performance; 

(2) Although ResNeXt has fewer parameters and GFLOPs than ResNet, the inference speed indicate that ResNeXt is much slower than ResNet, which means reducing FLOPs and model sizes does not always guarantee the reduction of GPU inference time and real energy consumption. In fact, ResNeXt has more MAC than ResNet (which has demonstrated in Eq.\ref{eq:eq4}), which may means the actual running speed is more related with MAC;

(3) Compared with the same layer VoVNetV2, TreeNet has less model complexity, similar GPU inference time, and higher accuracy. Specifically, the average decrease of parameters over VoVNetV2 is 34\% and the average performance increase is 0.63\% in terms of Top-1 accuracy, making TreeNet more suitable for real application scenarios. 

\begin{table*}[htb]
	\caption{Object detection results on COCO val2017. The best records and changes over VoVNetV2 are marked in \textbf{bold} and \textcolor[rgb]{ 0,  0,  1}{blue}, respectively.}
	\label{tab:tab4}
	\setlength{\tabcolsep}{5.5pt}{
		\begin{center}
			\begin{tabular}{c|c|cccccc}
				\toprule[1.2pt]
				Detectors & Backbones & AP50:95 & AP50 & AP75 & APs & APm & APl \\ 
				\midrule[1.1pt]
				\multirow{9}{*}{RetinaNet} & ResNet-50 & 36.5 & 55.4 & 39.1 & 20.4 & 40.3 & 48.1 \\ 
				\cmidrule{2-8}
				& VoVNetV2-39 & 37.0  & 56.8  & 39.8  & 20.8  & 40.7  & 48.1 \\ 
				\cmidrule{2-8}
				& \textbf{TreeNet-40} & \textbf{37.6} (\textcolor[rgb]{ 0,  0,  1}{$\uparrow0.6$})  & 57.3  & 40.1  & 22.2  & 41.0  & 49.4  \\ 
				\cmidrule{2-8}
				& ResNet-101 & 38.5  & 57.6  & 41.0  & 21.7  & 42.8  & 50.4  \\ 
				\cmidrule{2-8}
				& VoVNetV2-57 & 39.0  & 59.0  & 41.6  & 22.5  & 42.8  & 51.6 \\ 
				\cmidrule{2-8}
				& \textbf{TreeNet-58} & \textbf{39.6} (\textcolor[rgb]{ 0,  0,  1}{$\uparrow0.6$})  & 60.8  & 42.9  & 23.8  & 42.9  & 49.8  \\ 
				\midrule[1.1pt]
				
				\multirow{9}{*}{Faster RCNN} & ResNet-50 & 37.4  & 58.1  & 40.4  & 21.2  & 41.0  & 48.1  \\ 
				\cmidrule{2-8}
				& VoVNetV2-39 & 38.1  & 57.4  & 40.5  & 22.5  & 41.9  & 49.8 \\ 
				\cmidrule{2-8}
				& \textbf{TreeNet-40} & \textbf{38.8} (\textcolor[rgb]{ 0,  0,  1}{$\uparrow0.7$})  & 60.0  & 42.1  & 23.4  & 42.3  & 49.6  \\ 
				\cmidrule{2-8}
				& ResNet-101 & 39.4  & 60.1  & 43.1  & 22.4  & 43.7  & 51.1  \\ 
				\cmidrule{2-8}					
				& VoVNetV2-57 & 40.1  & 60.9  & 43.7  & 24.0  & 43.7  & 51.0 \\ 
				\cmidrule{2-8}
				& \textbf{TreeNet-58} & \textbf{40.9} (\textcolor[rgb]{ 0,  0,  1}{$\uparrow0.8$})  & 61.8  & 44.4  & 24.6  & 44.2  & 51.9  \\ 
				\bottomrule[1.2pt]
			\end{tabular}
	\end{center}}
\end{table*}

\subsection{Object Detection and Instance Segmentation on MS COCO}
We further apply TreeNet as backbones on the current state-of-the art detectors and segmenters, including RetinaNet~\cite{DBLP:conf/iccv/LinGGHD17}, Faster RCNN~\cite{DBLP:conf/nips/RenHGS15}, Mask RCNN~\cite{DBLP:conf/iccv/HeGDG17}, and MS RCNN~\cite{DBLP:conf/cvpr/HuangHGHW19}(all with feature pyramid(FPN~\cite{DBLP:conf/cvpr/LinDGHHB17}) structures). 

{\bf Settings.}
The experiments are conducted on COCO 2017, which contains 118k training, 5k validation and 20l test-dev images. We trained all models on the MS COCO train2017 split, and evaluated on the val2017 split.
We implement all detectors using the MMDetection toolkit\footnote{https://github.com/open-mmlab/mmdetection} with the default settings and trained them within 12 epochs(namely, ‘1× schedule’). For a fair comparison, we only replace the pre-trained backbones on ImageNet-1k and transfer them to MS COCO by fine-tuning, keeping the other components in the entire detector intact. 

\textbf{Results of object detection.} As shown in Table~\ref{tab:tab4}, detectors with both VoVNetV2 and TreeNet as backbones  outperform the original ResNet ones. Meanwhile, our TreeNet even gain a better performance than the VoVNetV2 version. Specifically, adopting the one-stage RetinaNet as the basic detector, the improvements are both 0.6\% in terms of box average precision(i.e., $\text{AP}^{box}$). If we employ Faster R-CNN as the base detector, the gains of $\text{AP}^{box}$ will be 0.7\% and 0.8\% for the 40-layers and 58-layers backbone models.

\begin{table}[htb]
	\caption{Instance segmentation results on COCO val2017. The best records are marked in \textbf{bold}.}
	\label{tab:tab5}
	\setlength{\tabcolsep}{16.0pt}{
		\begin{center}
			\begin{tabular}{c|c|c}
				\toprule[1.2pt]
				Segmentors & Backbones & AP50:95 \\
				\midrule[1.1pt]
				\multirow{6}{*}{Mask RCNN} & ResNet-50 & 34.7  \\
				\cmidrule{2-3}
				& \textbf{TreeNet-40 (Ours)} & \textbf{35.8}  \\
				\cmidrule{2-3}
				& ResNet-101 & 36.1  \\
				\cmidrule{2-3}
				& \textbf{TreeNet-58 (Ours)} & \textbf{37.5}  \\
				\midrule[1.1pt]
				
				\multirow{6}{*}{MS RCNN} & ResNet-50 & 36.0  \\ 
				\cmidrule{2-3}
				& \textbf{TreeNet-40 (Ours)} & \textbf{37.2} \\ 
				\cmidrule{2-3}
				& ResNet-101 & 37.6  \\ 
				\cmidrule{2-3}
				& \textbf{TreeNet-58 (Ours)} & \textbf{38.9}  \\
				\bottomrule[1.2pt]
			\end{tabular}
		\end{center}
	}
\end{table}

\textbf{Results of Instance Segmentation.}
As shown in Table~\ref{tab:tab5}, TreeNet achieves a clear improvement over the original ResNet, both on Mask RCNN and MS RCNN. Specifically, adopting Mask RCNN as the base segmentors, the gains are 1.1\% and 1.4\% in terms of $\text{AP}^{mask}$, respectively. Also, on MS RCNN, the improvements are 1.2\% and 1.3\%. These results verify our TreeNet has good generalization for various computer vision tasks.
 
\section{Conclusion}
In this paper, we propose an efficient yet effective network structure, named Tree block. Tree block extends the OSA module by replacing each $3\times3$ Conv layer with a stack of SRB and $1\times1$ Conv layer, where the $1\times1$ Conv layer is responsible for channel increasing and the SRB is fed into the next step. By doing this, the Tree block obtains less model complexity while achieving better performance over OSA. Based on the Tree block, we further build a lightweight backbone model TreeNet, which shares a similar structure to ResNet. Experimental results demonstrate TreeNet can significantly improve the performance of various computer vision tasks.

\bibliographystyle{splncs04}
\bibliography{ref}

\end{document}